\documentclass[letterpaper]{article} % DO NOT CHANGE THIS
\usepackage{aaai25}  % DO NOT CHANGE THIS
\usepackage{times}  % DO NOT CHANGE THIS
\usepackage{helvet}  % DO NOT CHANGE THIS
\usepackage{courier}  % DO NOT CHANGE THIS
\usepackage[hyphens]{url}  % DO NOT CHANGE THIS
\usepackage{graphicx} % DO NOT CHANGE THIS
\usepackage{natbib}  % DO NOT CHANGE THIS AND DO NOT ADD ANY OPTIONS TO IT
\usepackage{caption} % DO NOT CHANGE THIS AND DO NOT ADD ANY OPTIONS TO IT
\usepackage{subcaption}
\usepackage{algorithm}
\usepackage{algorithmic}

\usepackage{newfloat}
\usepackage{listings}
\DeclareCaptionStyle{ruled}{labelfont=normalfont,labelsep=colon,strut=off} % DO NOT CHANGE THIS
\floatstyle{ruled}
\newfloat{listing}{tb}{lst}{}
\floatname{listing}{Listing}
\title{EXPRESS: An LLM-Generated Explainable Property Valuation System with Neighbor Imputation}
\author {
    Wei-Wei Du*,
    Yung-Chien Wang*,
    Wen-Chih Peng
}
\affiliations {
    Department of Computer Science, National Yang Ming Chiao Tung University\\
    wwdu.cs10@nycu.edu.tw,~benwang.cs11@nycu.edu.tw,~wcpeng@cs.nycu.edu.tw
}

\usepackage{bibentry}
\newcommand\blfootnote[1]{
  \begingroup
  \renewcommand\thefootnote{}\footnote{#1}
  \addtocounter{footnote}{-1}
  \endgroup
}

\begin{document}

\maketitle
\blfootnote{$^*$Equal contribution.}
\begin{abstract}
The demand for property valuation has attracted significant attention from sellers, buyers, and customers applying for loans.
Reviews of existing approaches have revealed shortcomings in terms of not being able to handle missing value situations, as well as lacking interpretability, which means they cannot be used in real-world applications.
To address these challenges, we propose an LLM-Generated \textbf{EX}plainable \textbf{PR}op\textbf{E}rty valuation \textbf{S}y\textbf{S}tem with neighbor imputation called \textbf{EXPRESS}, which provides the customizable missing value imputation technique, and addresses the opaqueness of prediction by providing the feature-wise explanation generated by LLM.
The dynamic nearest neighbor search finds similar properties depending on different application scenarios by property configuration set by users (e.g., house age as criteria for the house in rural areas, and locations for buildings in urban areas).
Motivated by the human appraisal procedure, we generate feature-wise explanations to provide users with a more intuitive understanding of the prediction results\footnote{The source code is publicly available at https://github.com/benwangtch/express}.
% The source code\footnote{https://github.com/benwangtch/express} and demo video\footnote{https://youtu.be/W6X3tl0gbUc} are available.
\end{abstract}


\vspace{-5pt}
\section{Introduction}
Real estate appraisal is one of the most commonly used property valuation tasks.
In recent years, Automated Valuation Models (AVMs) have gained popularity in automating the valuation process \cite{mohd2020overview}.
These models are utilized not only by real estate buyers and sellers, but also by banks for mortgage lending purposes.
Various property valuation models, such as deep neural network models \cite{DBLP:conf/icdm/TanCW17, DBLP:journals/tist/LawPR19, DBLP:journals/tkde/PengLWYLZYH23, du2023dora, DBLP:journals/corr/abs-2302-00117, DBLP:conf/pakdd/LiWDP24}, have been proposed.
However, it is difficult to apply the above research to life, because users may not have access to all the details of the target property, leading to some missing values that cannot be incorporated into the model without imputation.
Imagine a user who wants to buy a house; how can the user know the Land Use Designation (e.g., Type A building land, Class B building site) for the target property?
Besides, current models offer little or no explainability of their predictions, and are therefore limited in their real-world applicability, as human appraisers find them challenging to trust.
% In addition to these models, several technology companies have developed their own online property valuation systems, including Zillow Zestimate \cite{zestimate}, HomeLight’s Home Estimates \cite{homelight}, and Chase Mortgage Services Home Value Estimator \cite{chase}.
% However, none of these systems is open source, nor do they disclose their technical details.
% Besides, property valuation highly depends on the regional factor and housing market, which makes it hard to directly apply one model all over the world.

To address these limitations, we developed \textbf{EXPRESS}, an open-source system that combines a reasonable imputation procedure \cite{8987530}, LLM-Generated feature-wise explanation, and an intuitive user interface.
In order to make the imputation procedure more flexible for different usages, the model incorporates a property configuration enabling users to specify filter conditions.
Then, the dynamic nearest neighbor search is employed to identify similar properties, followed by applying the neighbor imputation technique.
Since property valuation is a sensitive task that cannot rely on black-box models, we introduce the feature-wise explanation which is similar to the counterfactual explanation \cite{guidotti2022counterfactual} in the classification task.
Furthermore, we use an LLM to generate natural language explanations.
% To meet the needs of both professional and non-professional users, we developed an interface encompassing imputation results, property location visualization, and feature-wise explanation.
\begin{figure}
  \centering
  \includegraphics[width=\linewidth]{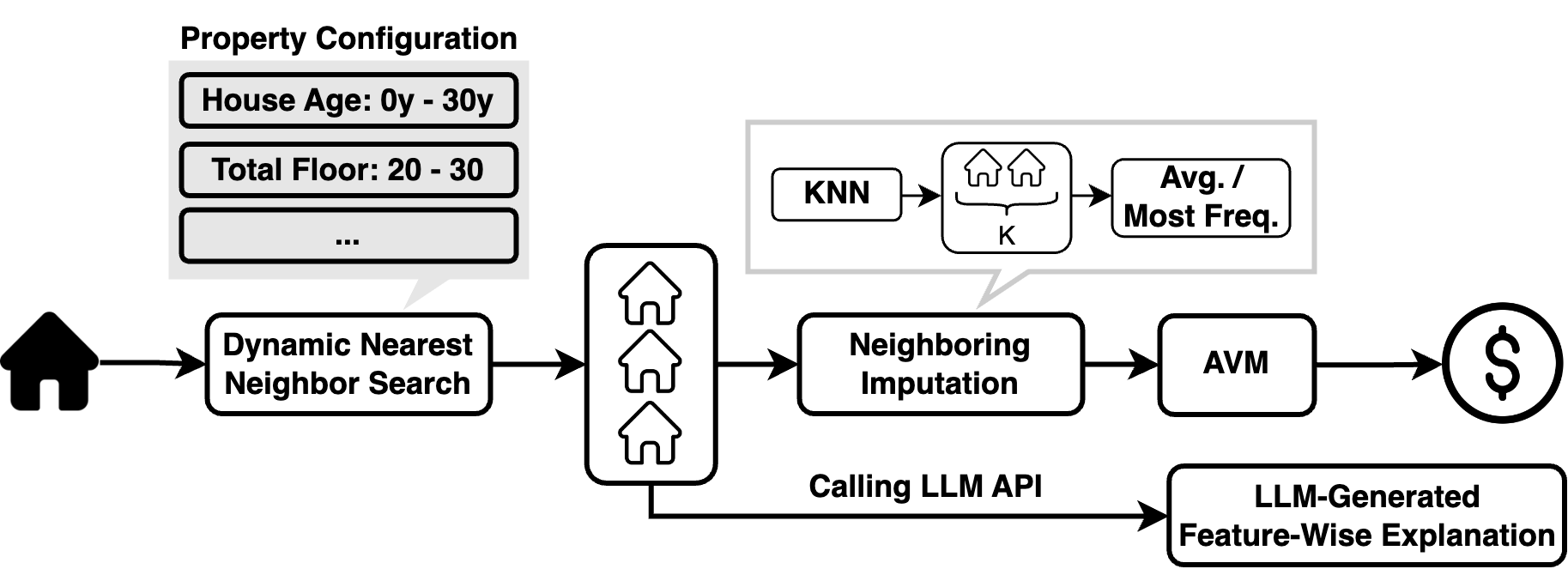}
  \caption{Property valuation process of our proposed system.}
  \label{fig:overview}
\end{figure}

\section{The EXPRESS framework}
Our datasets \cite{datasets} were collected from Taiwan, where property valuation needs to adhere to laws set up by the regulator, and the spatial relation of properties is significant.
% In Taiwan laws for property valuation, the valuation prediction process should compare with similar properties to ensure the reliability of the prediction, instead of using black-box model results.
% Inspired by the property valuation process of the human appraiser as shown in Figure \ref{fig:appraiser}, our system automates the entire procedure to improve efficiency and effectiveness while still retaining the concept of comparing with neighboring properties to comply with regulations.
As illustrated in Figure \ref{fig:overview}, our system comprises five modules: \textit{Property Configuration}, \textit{Dynamic Nearest Neighbor Search}, \textit{Neighbor Imputation}, \textit{Automated Valuation Module}, and \textit{LLM-Generated Feature-Wise Explanation}.

\subsection{Property Configuration}
The motivation of this module is to empower users with the capability to define neighboring conditions according to their specific requirements as shown in Figure \ref{fig:platform-1}.
For instance, professional appraisers who have a high degree of confidence regarding the target property can narrow the range to a more precise magnitude.
% In contrast, users with lower levels of certainty can set a broader range, so that the result will not be influenced by a limited subset of properties.
% The options of neighboring conditions, such as house age, main building area, floor area ratio, etc, are derived from domain knowledge and have demonstrated significance within gradient-boosted models.
Users can specify their desired numerical ranges for these features.
% For example, if users set the range of house age as five to twenty years, the system will exclusively consider properties falling within this temporal span as eligible neighbors.
This module enhances flexibility and customization for different application scenarios.

\subsection{Dynamic Nearest Neighbor Search}
The selection of similar properties is a crucial but time-consuming step in property valuation for human appraisers because they need to consider extensive domain knowledge and to search tens of thousands of properties.
In previous research, the k-Nearest Neighbors (kNN) algorithm can handle multiple missing values \cite{aljuaid2016proper} and offers flexibility in adjusting the number of neighbors.
As a result, we employ kNN to select similar properties.
% As a result, when the property filter provides a sufficient number of instances from the dataset, we employ the k-Nearest Neighbors (kNN) algorithm to select similar properties.
For the distance function, we employ the Minkowski distance, defined as follows:
\begin{equation}
    D(x, y) = \left( \sum_{i=1}^{n} |x_i - y_i|^2 \right)^{\frac{1}{2}}
\end{equation}
where $x_i$ is the i-th feature of the target property, $y_i$ is the i-th feature of the neighboring property, and n is the number of features.
% where $D(x,y)$ represents the Minkowski distance between the feature of the target $x$ and the feature of neighbor property $y$, while n is the number of features.

% Besides, the hyperparameter k is determined by the following equation:
% \begin{equation} \label{eq:k}
%     k = MIN(MAX(\#P_{neighbor}, 5), 20),
% \end{equation}
% where $\#P_{neighbor}$ is the number of neighboring properties selected by property configuration. 
% The value of k will dynamically vary between 5 and 20 depending on the input property and property configuration. 
% The lower bound of 5 is set to ensure users have adequate information, while the upper bound of 20 is set for two primary reasons. 
% It aims to prevent long inference times, as the search duration is linearly correlated with the value of k.
% For cases where the filtered properties are less than five, a hierarchical rule-based feature filtering module is utilized to search for similar properties.

% Additionally, since different property types have completely distinct characteristics \cite{du2023dora}, we have designed different filtering rules for each property type.

\subsection{Neighbor Imputation}
Accessing all the necessary features of the model is challenging for users, especially in property valuation tasks.
To address the situation of missing values, we employ the neighbor imputation based on the properties selected by dynamic nearest neighbor search.
Specifically, we average the values from neighboring properties for numerical features, employ the most frequent category for categorical features, and impute the most recent available value for time-dependent features.

\subsection{Automated Valuation Module (AVM)} 
This module was designed with flexibility in mind to accommodate the distinct characteristics of rural and urban areas, enabling the integration of various AVMs.
In our implementation, we apply LightGBM\cite{DBLP:conf/nips/KeMFWCMYL17}, which is a distributed and efficient gradient boosting framework that utilizes tree-based learning algorithms and has superior performance in most areas.
Apart from graph-based methods, which may be largely affected by biased neighbors, we consider the neighbor relation in previous modules and apply the tree-based method instead of neighbor-based methods in the price prediction stage.

\subsection{LLM-Generated Feature-Wise Explanation} 
\begin{figure}
  \vspace{-10pt}
  \centering
  \begin{subfigure}{0.5\textwidth}
          \centering
          \includegraphics[width=\linewidth]{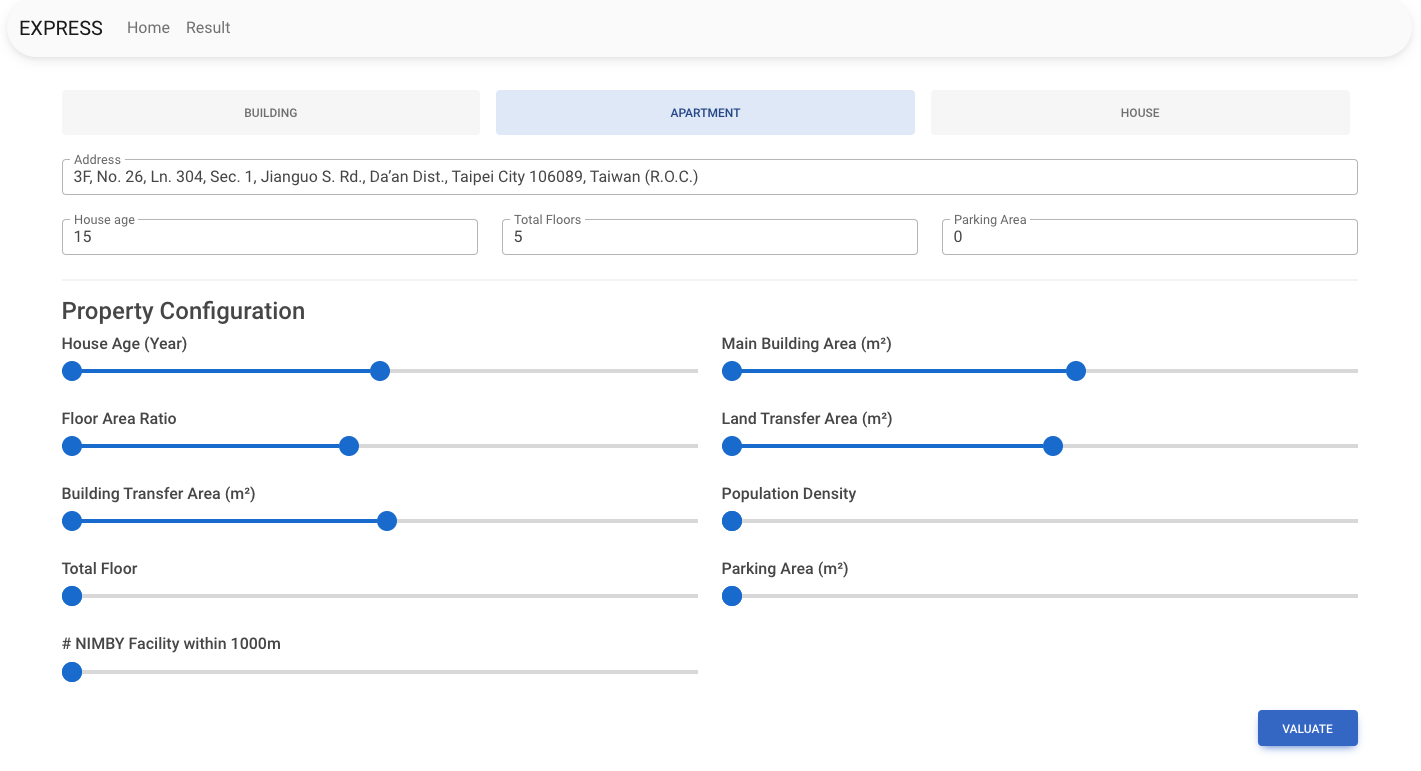}
            \vspace{-10pt}
          \caption{The interface of Property Configuration.}
          \label{fig:platform-1}
  \end{subfigure}
  \begin{subfigure}{0.5\textwidth}
          \centering
          \includegraphics[width=\linewidth]{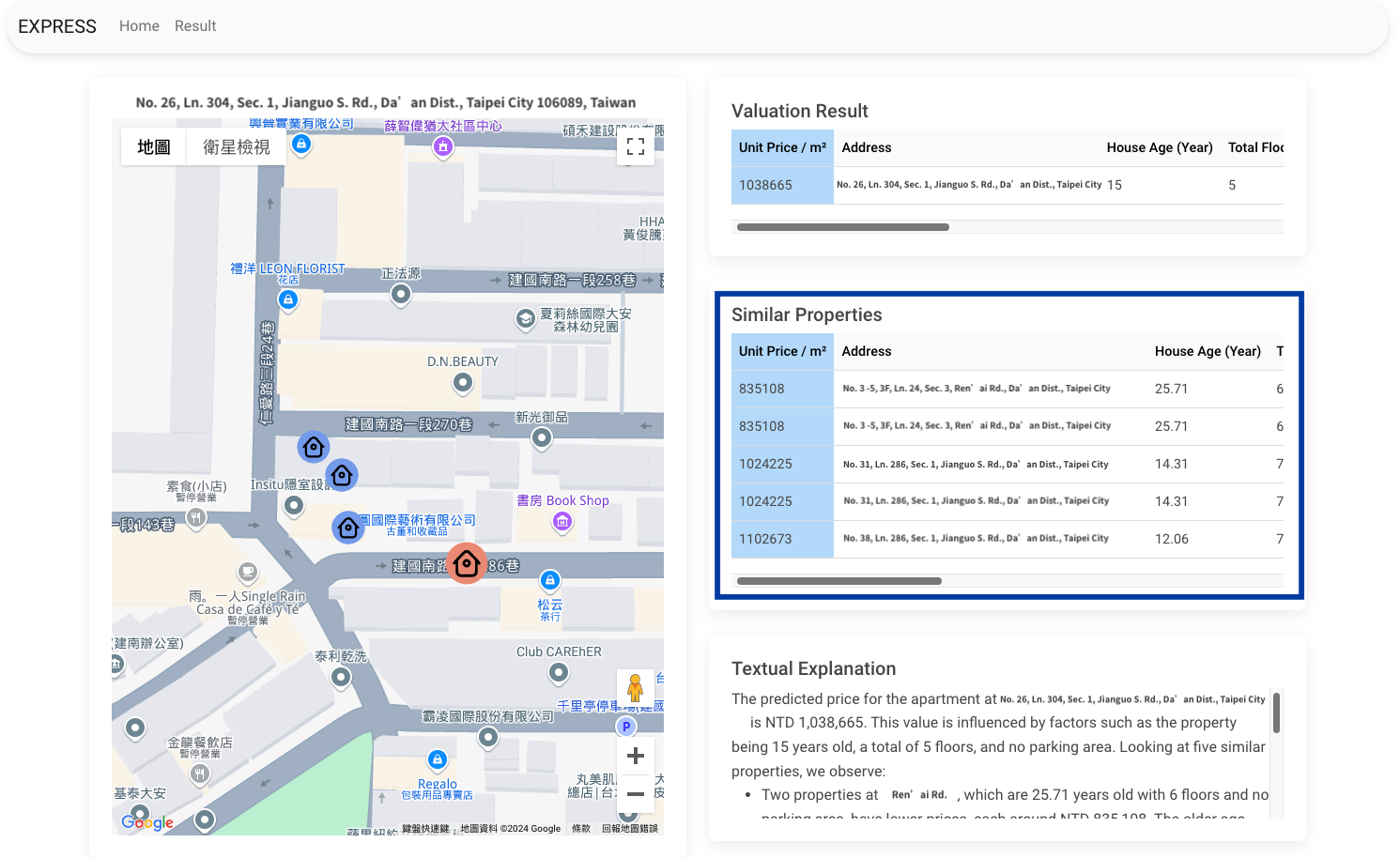}
          \caption{The interface of Prediction Result.}
          \label{fig:platform-2}
  \hfill
  \end{subfigure}
   \vspace{-20pt}
  \caption{System interface of EXPRESS.}
 \vspace{-10pt}
\end{figure}

Through dynamic nearest neighbor search, as shown in the blue frame in  Figure \ref{fig:platform-2}, we utilize similar properties and a large language model (LLM) to provide interpretability for the target predicted property.
Previous studies in recommender systems have combined past preferences with new predictions to generate explanations using language models, through the design of prompting strategies \cite{lubos2024llm}.
Inspired by this approach, we design prompts and use an LLM to generate feature-wise explanations by comparing the differences in features between the target property and similar properties.
Both the target predicted property and the similar properties in JSON format \cite{hegselmann2023tabllm} are input into the language model, enabling it to deliver feature-wise and comprehensible explanations of the prediction for users.

% The user interface presents all the features and unit prices of both the target property and its neighbor properties.
% Users can independently evaluate the unit price estimated by the AVM using these feature-wise explanations.

\vspace{-5pt}
\section{Conclusion}
EXPRESS is an open-source system with neighbor imputation and LLM-generated feature-wise explanations for property valuation tasks.
It solves the issue of missing values, which is challenging for users who may not have access to all the features required for real-world property valuation models.
Specifically, we introduce the dynamic nearest neighbor search to find neighboring properties based on the property configuration set by users, and then impute the missing value features by the neighboring properties.
In order to comply with the law, the LLM-generated feature-wise explanation is provided to offer users a deeper understanding of the valuation results, unlike the black-box prediction result.
The explanation not only improves transparency but also empowers users to make their own judgments based on the provided natural language explanation.
% Finally, an easy-to-use UI was designed after user study with both non-domain users and professional appraisers in financial institutions; it was designed according to their actual use procedures.
% According to our interview with appraisers in the bank, human appraisals in Taiwan take half a day per property. 
% Our system significantly reduced the appraisal time to five minutes.
% In the future, the model framework can be extended to support any additional features and models.
Different from existing black-box online property valuation services, our aim with EXPRESS is to address the real needs of users, to support users' participation in the valuation process, and to facilitate research growth in property valuation.

\appendix
% \section{Reference Examples}
\label{sec:reference_examples}

\bibliography{aaai25}

\end{document}